\documentclass[final,1p,times]{elsarticle}
\usepackage{amsmath}
\usepackage{amssymb}
\usepackage{lineno,hyperref}
\usepackage{algorithm}
\usepackage{algorithmic}
\usepackage{graphicx}
\usepackage{multirow}
\usepackage{colortbl}

\journal{Journal of \LaTeX\ Templates}
\newcommand{\cl}{\mathcal}
\newcommand{\bs}{\boldsymbol}
\newcommand{\etal}{\emph{et al.}}
\newcommand{\mycite}{\cite}
\newcommand{\ie}{\emph{i.e.}}
\newcommand{\eg}{\emph{e.g.}}
\DeclareMathOperator*{\argmin}{\arg\!\min}
\definecolor{Blue}{rgb}{0.5176,0.4392,1.0000}
\newcolumntype{g}{>{\columncolor{Blue}}c}

\def\bestviewed{\textit{Best viewed in color.}}

\newcommand{\revise}[1]
{
	{\color{black}{#1}}
}

\newcommand{\pos}{\bs x}
\newcommand{\numfeat}{N}

\newcommand{\feat}{\bs f}
\newcommand{\nomenclature}[1]{}

\def\urlispg{\url{http://sites.uclouvain.be/ispgroup/index.php/Softwares/HomePage}}

\def\emails{\{amit.kc, christophe.devleeschouwer\}@uclouvain.be, damien.delannay@keemotion.com}
\def\affiliationakc{ISPGroup, ELEN Department, Universit\'{e} catholique de Louvain, Belgium}
\def\affiliationdd{Keemotion, Belgium}

\begin{document}

\begin{frontmatter}

\title{Iterative hypothesis testing for multi-object tracking in presence of features with variable reliability}

\tnotetext[mytitlenote]{AKC and CDV are funded by the Belgian National Science Foundation (FRS-FNRS).}

\author{Amit Kumar K.C.$^{1}$, Damien Delannay$^{2}$ and Christophe De Vleeschouwer$^{1}$}
\address{$^{1}$ \affiliationakc \\ $^{2}$ \affiliationdd \\ \emails}

%
%
%
\begin{abstract}
This paper assumes prior detections of multiple
targets at each time instant, and uses a graph-based approach to connect
those detections across time, based on their position and appearance
estimates. In contrast to most earlier works in the field, our framework
has been designed to exploit the appearance features, even when they are
only sporadically available, or affected by a non-stationary noise, along
the sequence of detections. This is done by implementing an iterative
hypothesis testing strategy to progressively aggregate the detections into
short trajectories, named tracklets. Specifically, each iteration considers
a node, named key-node, and investigates how to link this key-node with
other nodes in its neighborhood, under the assumption that the target
appearance is defined by the key-node appearance estimate. This is done
through shortest path computation in a temporal neighborhood of the
key-node. The approach is conservative in that it only aggregates the
shortest paths that are sufficiently better compared to alternative paths.
It is also multi-scale in that the size of the investigated neighborhood is
increased proportionally to the number of detections already aggregated
into the key-node. The multi-scale nature of the process and the progressive relaxation of its conservativeness 
makes it both computationally efficient and effective.

Experimental validations are performed extensively on a toy example, a 15 minutes long
multi-view basketball dataset, and other monocular pedestrian datasets.
\end{abstract}
\end{frontmatter}

\section{Introduction} 
\label{section:introduction_iht}
Multi-object tracking (MOT) is a fundamental issue in computer vision. It supports high-level semantic scene analysis in numerous and various applications. Vehicle trajectories are, for example, collected to control traffic monitoring solutions \cite{video_tracker_traffic_monitoring}. People displacement analysis is important to improve the security of public spaces \cite{anomalous_event_detection}, or to understand sport actions \cite{sparsity_sport_player_detection_tracking}. In microscopy, tracking of cells helps to understand biological processes \cite{kaakinen2014automatic}.

\subsection{Detection-based MOT problem formulation}
\label{section:mot_formulation}
Due to recent improvements in object detection, many detection-based approaches have been proposed to handle the MOT problem. In such approaches, plausible object locations are first estimated in each individual frame and some features, characterizing the appearances of the detected objects, are extracted. Afterwards, the MOT problem is formulated as the problem of grouping these detections into a minimum number of disjoint trajectories, each trajectory corresponding to a single physical entity. This \emph{data association problem} is usually handled by graph-based solutions. First, a graph is defined to connect a set of nodes that correspond to the detections (or unambiguous association of detections, named \textit{tracklets}). Each edge gets a weight that reflects either distance (or dissimilarity) or similarity in terms of spatio-temporal displacement and/or appearance between the two nodes it connects. Afterwards, multi-object tracking can be formulated in its general form as the problem of partitioning the graph into disjoint sets $T_i, i>0$ of nodes \cite{gmcp_tracker} such that
\begin{itemize}
	\item each set contains one and only one detection at each time instant\footnote{Potential missed detections (MD) or appearing/vanishing targets are typically handled based on virtual nodes. The inclusion of such a virtual node is a set $T_i$ induces a penalty to avoid selecting the virtual node option if the frame includes a proper detection.},
	\item each detection is included in one and only one of the sets\footnote{False positive detections (FP) are typically handled by defining a false detection set $T_{FP}$ that gathers all detections that are not part of a trajectory set $T_i$. A false positive penalty is assigned to each element in $T_{FP}$ to avoid the inclusion of correct detections to this set. For simplification purposes, we ignore MD and FP in the rest of this introduction section.}
	\item the elements of a set are consistent in terms of appearance and spatio-temporal features, and
\end{itemize}

Formally, this can be written as
\begin{align}
	\begin{array}{ll}
		\text{minimize} & {\sum_{i=1}^{K}{C(T_i)}}, \\
		\text{subject to} & {T_i  \cap T_j =\emptyset , \forall i \neq j},\\
		{} & 				{\cup_{i=1}^K T_i=\cl V}, \\
		{} & \forall i>0, \text{ and } \forall u,v \in T_i: t_u \neq t_v, \\
		{} & \forall i>0, \text{ and } \forall t, \exists u \in T_i \text{ with } t_u=t.
	\end{array}
	\label{eqn:overall_objective}
\end{align}
where $\cl V$ represents the set of all nodes, 
$C(T_i)$ represents the dissimilarity cost within the $i$-th set $T_i$, and $t_u$ represents  the associated time of node $u$. In Equation~(\ref{eqn:overall_objective}),  the first two constraints require that the sets $\{T_i\}_{i=1}^K$ define a valid partition, whereas the last constraint requires that $T_i$ cannot have multiple detections from the same time instant. The cost $C(T_i)$ should be defined such that it decreases (increases) when the detections in $T_i$ have a small (large) dissimilarity between them, reflecting (in)consistent associations. The quality of the solution relies on the definition of $C(T_i)$. Ideally, if there are $n$ nodes within $T_i$, the dissimilarity function should consider all $n(n-1)/2$ of time-causal pairs of nodes, and associate to each of them a cost that increases with the likelihood that the nodes in a pair correspond to two distinct targets. That is, 
\begin{equation}
	C(T_i):=\sum_{\substack{u,v \in T_i \\ u \neq v}} w_{uv},
	\label{eqn:generic_definition_cost_track}
\end{equation}
where $w_{uv}$ is defined to decrease with the likelihood that the nodes $u$ and $v$ correspond to the same physical object in terms of space, time and/or appearance. Typically, $w_{uv}$ should increase with the appearance dissimilarity and the spatial distance between nodes $u$ and $v$. More importantly, its definition should also account 
\begin{enumerate}
	\item  \label{observation1_iht} for the time elapsed between $u$ and $v$: a larger time interval makes it more likely that a target has moved or changed in appearance, hence reducing $w_{uv}$ for a given observed dissimilarity
	\item \label{observation2_iht} for the confidence we have in the observations: an unreliable feature should not lead to definitive conclusion about whether the nodes correspond to the same target or not.
\end{enumerate}

In the following, we refer to these two principles as \emph{time evanescence} and \emph{reliable feature prominence} respectively.

\subsection{Previous art simplification and related issues}
\label{section::previous_art_simplification_iht}
Given the definition of $C(T_i)$, provided in Equation~(\ref{eqn:generic_definition_cost_track}), solving Equation (\ref{eqn:overall_objective}) rapidly becomes computationally intractable. As pointed out by Zamir \etal \mycite{gmcp_tracker}, the problem becomes equivalent to the travelling salesman problem, which is known to be NP-complete. Therefore, most previous works build on the time evanescence principle described above, namely on the fact that $w_{uv}$ should only be large for nodes that are close in time, to simplify the problem. Specifically, they ignore dissimilarities between far away nodes and only consider for each node $u$ the cost $w_{uv^\star}$ induced by its immediately subsequent node $v^\star$ in $T_i$. Formally,

\begin{align}
	C(T_i):=\sum_{u, v^\star \in T_i}w_{uv^\star},
	\label{eqn:uv_star}
\end{align}
where $v^\star:=\argmin_{v \in T_i, t_v>t_u} (t_v-t_u)$ is the node in $T_i$ that is temporally the closest to $u$.

Doing so, Equation~(\ref{eqn:overall_objective}) becomes easy to solve, since it basically reduces to finding a set of paths with a minimal cumulative cost. This can be solved by using a greedy shortest-paths computation \cite{dijkstra1959note}, or by running the K-shortest paths (KSP) algorithm \cite{berclaz2011ksp}. Apart from the KSP, several other algorithms such as network flow algorithm \cite{zhang2008global}, robust hierarchical association \cite{huang2008robust} can be envisioned to estimate the $K$ tracks under this simplification assumption. These approaches have been proven to be effective in a variety of scenarios for which the prominence of the links connecting close observations is valid in many practical association problems.

This simplification, however, fails to correctly model the tracking problem when the cost $w_{uv}$ of the links that connect nodes that are distant in time becomes important compared to the links between subsequent observations. This typically happens when discriminant features are observed with variable level of reliability along the time. 
In this case, due to the reliable features prominence principle, the cost $w_{uv}$ becomes relatively more significant (either smaller or larger depending on whether the nodes are similar or not) between far away, but reliably observed, nodes than between close nodes with noisy features. Such cases are prevalent in numerous practical scenarios. For example, color histograms appear to be quite noisy in presence of occlusions, and in some other cases, highly discriminant appearance features are only available sporadically (and under certain configurations only). 
For example, in sports, a number on a jersey is visible only when facing the camera. 

\begin{figure}[t]
	\centering
	\includegraphics[width=0.9\linewidth]{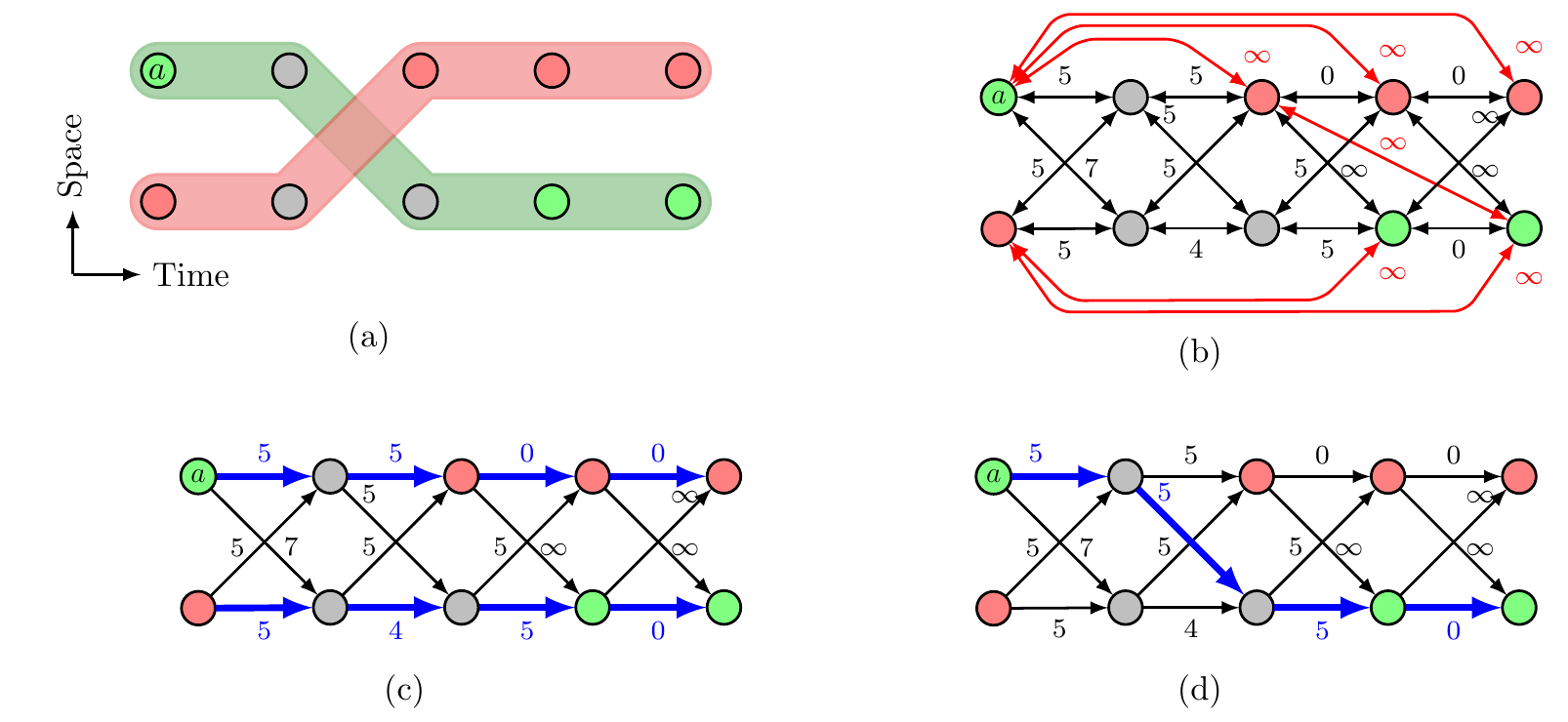}
	\caption[Problem of conventional tracking method in presence of sporadic appearance features]{\textbf{Problem of conventional tracking method in presence of sporadic appearance features.} \textbf{(a)} Detections and trajectories corresponding to two targets (red and green) for 5 consecutive frames are shown. Gray nodes do not have appearance features. For readability, \textbf{(b)} only depicts a subset of edges of the fully connected graph considered in Equation (\ref{eqn:generic_definition_cost_track}) (see text for details). \textbf{(c)} Conventional tracking algorithms, \ie, with the time-evanescence simplification assumption, fail to track the target correctly, and result in appearance inconsistencies along a track. \textbf{(d)} Given the appearance of the key-node $a$, it is possible to simply increase (respectively, decrease) the cost of going through the nodes that are dissimilar (respectively, similar) in the graph irrespective of whether the nodes are temporally close or far. The resulting shortest-path, shown by thick blue arrow, from $a$ is consistent with all the available appearances. \textit{ Best viewed in color.}}
	\label{fig:problem_conventional_sporadic}
\end{figure}

In such time-varying observation processes, the task of tracking multiple objects, while taking into account the position and all the available appearance features, cannot be addressed properly with the formulation in Equation~(\ref{eqn:uv_star}). This is due to the fact that the consistency of a track cannot be measured by the mere accumulation of (dis)similarities between the consecutive nodes in the track, simply because the appearance features might be unreliable or even purely unavailable in some nodes. This major shortcoming of conventional graph-based tracking is illustrated in Figure~\ref{fig:problem_conventional_sporadic}.

Figure~\ref{fig:problem_conventional_sporadic}(a) depicts the ground-truth trajectories of a red and a green target, as well as the appearance observed in each time frame for each of the target. The color of the node indicates whether the color of the target is available (red or green) or unavailable/unreliable (gray). 
The problem, defined by Equations~(\ref{eqn:overall_objective}) and (\ref{eqn:generic_definition_cost_track}), is depicted in Figure~\ref{fig:problem_conventional_sporadic}(b). Edge cost is zero when connecting two nodes with the same color, intermediate (and function of spatio-temporal measurements ) when the color information is lacking for one of the nodes, and infinite when the connected nodes have distinct colors. For readability, only the edges connecting the detections that are observed at consecutive times are depicted (in black), plus the edges with infinite weight (in red). Other $w_{uv}$ are negligible due to the fact that $w_{uv}$ has to decrease as time elapses between $u$ and $v$ (time evanescence principle discussed above). The solution to problem (\ref{eqn:overall_objective}), computed from this graph, using exhaustive search approach, corresponds to the desired tracks and is depicted in Figure~\ref{fig:problem_conventional_sporadic}(a). In contrast, making the simplification assumption presented in ~(\ref{eqn:uv_star}) and thus omitting all links between non-consecutive nodes, fails to track the target correctly. This is depicted in Figure~\ref{fig:problem_conventional_sporadic}(c), where we observe that a conventional ($K$-)shortest approach ends up in associating red and green nodes.

Interestingly, we also observe from this toy-example that, if we were specifically interested in tracking the green target observed in the node $a$ depicted on the top left of Figure~\ref{fig:problem_conventional_sporadic}(d), a trivial solution would consist in increasing/decreasing the cost of an edge when it enters a red/green node, wherever they occur along the track. In that way, the shortest-path to connect node $a$ to the window extremity would become consistent with the color observations. 

In this paper, as a primary contribution, we propose to extend this trivial single-target tracking solution to a multi-object tracking context, in which no prior knowledge is available about the actual appearance of the targets, and in which the appearance measurements are subject to noise that changes over time (non-stationary), but whose relative importance is known as a prior. In practice, this prior is typically derived from the detector (which might reveal an occlusion that hampers the appearance observation) or from the feature measurement process (\eg, a digit recognition algorithm might conclude that no digit is visible or that its recognition is quite uncertain).

\subsection{Contribution}
\label{section::contribution_iht}
To circumvent the limitations of conventional algorithms, we propose a new paradigm to aggregate detections into objects trajectories. It extends the trivial solution depicted in Figure~\ref{fig:problem_conventional_sporadic}(d) by emebdding shortest-paths computations within an \emph{iterative hypothesis testing} (IHT) strategy. 

Each iteration of the algorithm works as follows. A node, named key-node (node $a$ in Figure~\ref{fig:problem_conventional_sporadic}(d)), is selected to define a target appearance hypothesis. Given this hypothesis, a shortest-path algorithm is considered to investigate how to aggregate the key-node with its temporal neighbors in the graph, while promoting the nodes that share its appearance, just as for node $a$ in Figure~\ref{fig:problem_conventional_sporadic}(d). The process is repeated iteratively, each node possibly becoming a key-node at some step of the algorithm.
To avoid misleading the overall multi-object tracking process due to a wrong intermediate aggregation decision, \eg, caused by some inappropriate appearance hypothesis, the shortest-path connecting the key-node to its neighborhood is only validated  when it is `sufficiently shorter' than alternative paths. 
The criterion to validate the shortest-path is very strict in the beginning of the iterative process but is then progressively relaxed as the iterations proceed. This progressive relaxation makes the process \emph{greedy} in the sense that most reliable tracklets will be extracted first, independently of the order in which nodes are scheduled as key-nodes. 

Another worthwhile design choice consists in adapting the observation window to the size of the key-node (\ie, number of detections already aggregated into the key-node), making the process \emph{multi-scale}. The advantages are two-fold. First, it reduces complexity by aggregating the nodes locally before considering larger observation windows. Second, it gives the opportunity to investigate long time horizons based on more reliable appearance information (since appearance has been accumulated on many frames for large key-nodes), which benefits the tracking accuracy.

The proposed approach has the advantage of naturally accounting for different levels of reliability in the observation process, typically by giving more credit to the reliable appearance measurements when defining the cost associated to the discrepancy between the target appearance hypothesis and a node appearance estimate. Hence, the algorithm becomes able to effectively exploit sporadic features or features whose reliability varies over time, which is a significant step forward compared to the state-of-the-art. 

Compared to our initial work presented in \cite{kc2012iterative}, this journal paper:
\begin{itemize}
	\item positions our algorithm with respect to the generic MOT formulation defined by Equations (\ref{eqn:overall_objective}) and (\ref{eqn:generic_definition_cost_track}), revealing that a larger range of practical problems can be addressed our solution, compared to the ones supported by conventional simplification adopted in Equation (\ref{eqn:uv_star}),
	\item introduces the progressive relaxation concept, which is a valuable extension compared to our conference paper since it avoids the tedious and hazardous tuning of the thresholds associated to the validation criteria (see the testing phase in Section \ref{section:ambiguity_estimation_and_validation}),
	\item extends the off-line algorithm presented in \cite{kc2012iterative} to on-line tracking scenarios (see Section \ref{section:incremental_iht}),
	\item releases a public reference software implementation of our algorithm\footnote{\urlispg},
	\item provides extensive validations both on synthetic and real-life data, which helps in assessing the practical usability and relevance of our approach.
\end{itemize}

The rest of the paper is organized as follows. Section \ref{section:related_works_iht} presents the (few) methods that have been previously introduced to handle sporadic features or features with time-varying reliability. Section~\ref{section:graph_formalism_and_notations} defines the graph terminology. Our iterative hypothesis testing algorithm is described and discussed in Section~\ref{section:iht_global_algorithm}. Section \ref{section:incremental_iht} extends our approach to on-the-fly incremental tracking scenarios. Section~\ref{section:evaluation_iht} presents the experimental results, and demonstrates the efficiency and effectiveness of our approach both on a synthetic and a real-life datasets.
\section{Related works}
\label{section:related_works_iht}
In this section, we review the few works that have been proposed to address the multi-object tracking in presence of features that are sporadic and/or affected by non-stationary noise.

The global appearance constraint (GAC) approach \mycite{shitrit2011global} assumes a prior knowledge of a discrete set of $N$ possible appearances, and ends up in computing $K$-shortest paths in a $N$-layered graph, $K$ being the number of targets, and $N$ corresponding to the number of possible target appearances. In contrast, to avoid the computational burden associated to the construction of a $N$-layer graph, and to handle cases for which the possible set of appearances is not a known and finite discrete set, we embed the hypothesis testing within an iterative local aggregation framework. We show in our validation that this results in significant accuracy improvements. 

The discriminative label propagation (DLP) approach \cite{kc2013discriminative} presents an elegant method to combine various appearance and spatio-temporal relationships between the detections by constructing a number of complementary graphs, and assigning labels to these detections in a manner that is consistent with all graphs. This approach only handles sporadic appearance features, and can thus not take advantage of continuous features reliability priors.



Zamir \etal \mycite{gmcp_tracker} adopt a similar formulation than the one defined in Equation~\ref{eqn:generic_definition_cost_track} but apply it on short (typically 50 frames long) segments, for computational tractability. The solution on each segment is obtained by solving a generalized minimum clique problem (GMCP) based on a greedy heuristic. 
The same procedure is repeated in a hierarchical manner to generate long trajectories. In addition to the sub-optimality of the GMCP solution, a drawback of this approach lies in the fact that the decision have to be taken at each level of the hierarchy before moving to the next level. As a consequence, lower level decisions (derived from only partial information) might be wrong and impact the final solution. In contrast, our approach works conservatively and does not force decisions on small observation windows when those decisions are ambiguous (testing phase in Section \ref{section:ambiguity_estimation_and_validation}). 

\section{Graph formalism and notations}
\label{section:graph_formalism_and_notations}
As an input, the algorithm receives the set of candidate targets, detected independently at each time instant as described in \mycite{delannay2009detection}. Apart from the detection time $t$ and the location
$\pos$, the detector computes $\numfeat$  appearance features $\bs \feat_i$ ($1\leq i\leq \numfeat$) for a target. Since a
feature might be unreliable or even missing, the detector outputs a
confidence value $c_i \in [0, 1]$ for each feature ($c_i=0$ standing
for a missing feature).  A detection $\bs d$ is therefore
characterized by the vector
\begin{equation*}
\bs d = (t,\bs y, \cl F, \bs c),
\end{equation*}
where $\cl F = \{\bs f_1,\,\cdots, \bs f_{\numfeat}\}$ and $\bs c=(c_1,\,\cdots,
c_{\numfeat})$. The set of detections at a given time $t$ is denoted as
$\mathcal{D}^t$.  
As introduced earlier, the proposed algorithm adopts a graph-based formalism. 
We define a graph $\mathcal{G}=\left(\mathcal{V},\mathcal{E},\bs{W}\right)$ by:
\begin{itemize}
\item a set of nodes, with each node corresponding to a tracklet, \ie, \\ $\mathcal{V}=\{v_k|1\leq k\leq |\cl V|\}$,
\item a set of edges, $\mathcal{E} \subset\mathcal{V} \times \mathcal{V}$, defining the connectivity between the nodes in $\mathcal{V}$,
\item and a set of weights, $\bs{W}: \mathcal{E}\rightarrow
      \mathbb{R}_{+}$, weighting these nodes and edges.
\end{itemize}
Initially, individual detections define the nodes of the graph. Detections are then aggregated into \emph{tracklets}, which define the nodes of the updated graph. The proposed iterative aggregation process is presented in details in Section \ref{subsection:iterative_tracklet_aggregation}, including the definition of cost and edges between nodes. Here, we only introduce the associated terminology. Formally, a tracklet $v$ is defined to be collection of chained detections, \ie, $v=\left(\bs d^{1},\bs d^{2},\,\cdots, \bs d^{|v|}\right)$, $|v|$ being the length of
the tracklet. Notice that the chain is ordered in the sense that the detection
times $t_{d^{(i)}}$, $i \in [1,|v|]$ are such that $t_v^{(s)}=t_{d^{1}}<t_{d^{2}}<\cdots<t_{d^{|v|}}=t_v^{(e)}$, with $t_v^{(s)}$ and $t_v^{(e)}$ respectively denoting the starting and ending time of the tracklet.

Notice that pairs of tracklets are connected only between their extremities in
a way that maintains the increasing ordering of the detection times composing the two tracklets.  The weight $w_{uv}$ is introduced to denote the linking cost between two nodes $u, v\in \cl V$. It is formally defined in Section~\ref{section:graph_construction_iht}. In short, it typically decreases with the likelihood that the nodes $u$ and $v$ correspond to the same physical target. In addition, we introduce the \emph{inner cost} $w_{v}$\footnote{Note that $w_{v}$ is not the self-loop of $v$.} of a node $v$ to denote the cost of traversing tracklet $v$ from its starting time to its ending time. It is introduced to avoid that long nodes create \emph{short-cuts} in the graph. Since the edges are directed and ``time-forwarded'' (see Section~\ref{section:graph_construction_iht}), the graph $\mathcal{G}$ is directed and acyclic (DAG)
, and permits only causal traversals. Nevertheless, the graph can be globally reversed in order to allow anti-causal paths for processing purposes. We denote such reversed graph as $\cl G^{-}$. 

In the sequel, we use two more graph notations. First, $\cl G_{\delta}$ represents a windowed-graph formed by
selecting in $\cl G=(\cl V, \cl E, \bs W)$ the tracklets $v \in \cl V$
having at least one extreme time component inside the temporal window
$\delta$. The connectivity $\cl E$ and the weight $\bs W$ are
restricted accordingly from these selected tracklets in order to form
$\cl E_{\delta}$ and $\bs W_{\delta}$ respectively. Second, in case of incremental tracking, the algorithm incorporates new detections at each time instant $t$ and the graph is continuously incremented with time. We denote the graph at time $t$ by $\cl G^t$. The corresponding vertices and edges are denoted by $\cl V^t$ and $\cl E^t$ respectively.

\begin{figure}[t]
	\centering
	\includegraphics[width=0.5\linewidth]{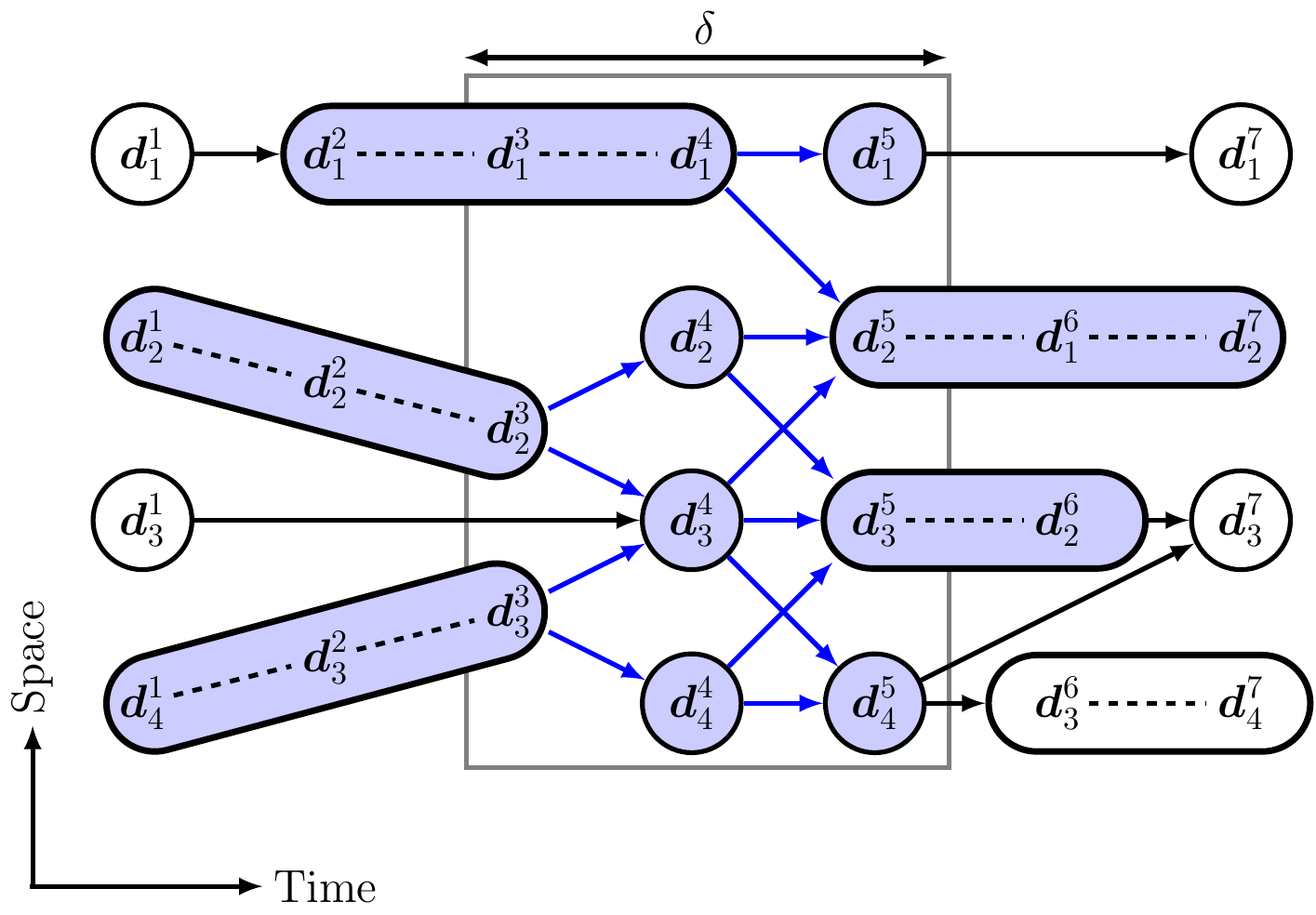}
	\caption{\textbf{Graph formalism for iterative hypothesis testing.} The $k$-th detection at time $t$ is denoted by $\bs d_k^t$. They are aggregated into tracklets. Each node corresponds to a tracklet. An edge that connects two nodes $u$ and $v$ has a cost $w_{uv}$. The windowed-graph $\cl G_{\delta}$ is comprised of the nodes and (blue) edgeswithin the observation window $\delta$.}
	\label{Fig:graph_representation}
\end{figure}

Figure~\ref{Fig:graph_representation} depicts how the tracklets are gathered into a graph in the proposed framework. 
\section{Iterative hypothesis testing algorithm}
\label{section:iht_global_algorithm}
This section first explains the construction of graph. Afterwards, it presents our proposed algorithm, and outlines its characteristics.

\subsection{Graph construction}
\label{section:graph_construction_iht}
As introduced earlier, our nodes correspond to tracklets. We create a directed edge from $u$ to $v$ only if $0 < t_v^{(s)} - t_u^{(e)} \leq \tau_{\rm max}$, \ie, node $v$ occurs after $u$ and the time interval is smaller than $\tau_{\rm max}$. The weight of the edge $w_{uv}$ is defined solely by the spatio-temporal displacement between $u$ and $v$, \ie,

\begin{align}
	w_{uv}:=\left\{
	\begin{array}{ll}
		[1+\gamma\cdot(t_v^{(s)}-t_u^{(e)}-1)] g_{\rm sp}(u,v) & \text{ if } 0 < t_v^{(s)} - t_u^{(e)} \leq \tau_{\rm max}, \\
		\infty & \text{ otherwise }
	\end{array}
	\right.
	\label{eqn:spatio_temporal_cost}
\end{align}

where the factor $\gamma>0$ introduces penalty for missed detections, and $g_{\rm sp}(u,v)$ measures the distance between $v$ and the predicted position of the object corresponding to node $u$. It is defined as
\begin{equation}	
	g_{\rm sp}(u,v):=\big\|\bs{y}^{(s)}_v- \bs{y}^{(e)}_u -\dot{\bs{y}}^{(e)}_u\big(t_v^{(s)}-t^{(e)}_u\big)\big\|_2,
\end{equation}
where the term $\dot{\bs{y}}^{(e)}_u$ is the velocity, at the end of tracklet $u$. It is zero for unit length tracklets, and is computed from the last 2 detections of the tracklet otherwise. Since the edges are directed and ``time-forwarded'', the graph $\mathcal{G}$ is directed and acyclic (DAG).
\subsection{Iterative hypothesis testing}
\label{subsection:iterative_tracklet_aggregation}
Our major objective is to design a detections aggregation method that is able to exploit appearance cues even when they are sporadic or have variable reliability. Therefore, we promote a novel paradigm, founded on an  \emph{iterative hypothesis testing} process. 

\textbf{Overview of the contribution}
In this approach, each iteration selects a node, named key-node, and computes the shortest-path to connect this key-node to the extremity of either a forward or backward neighborhood, under the assumption that the observed key-node appearance defines the reference appearance of the tracked object. Given this hypothesis, paths that go through nodes that do (not) share the key-node appearance are promoted (penalized). This is done simply by decreasing (increasing) the cost to go through a node of the graph when the appearance of that node is similar (different) to that of the key-node. Hence, all appearance cues, even the sparse or inaccurate/unreliable ones, can be exploited to drive the selection of aggregated paths within the graph. Since the process is repeated with each node being the key-node, all observed appearance hypotheses are examined. 

Two subtle mechanisms largely contribute to the success of our approach:
\begin{itemize}
	\item The first and primary one lies in the conservativeness adopted to turn the key-node shortest-path into a single tracklet node for subsequent iterations. Actually, this path is only validated if it is sufficiently better than alternative paths. Importantly, the notion of `sufficiently good', which is formally defined in Section \ref{section:ambiguity_estimation_and_validation} below, is progressively relaxed along the iterative process. This makes the overall algorithm greedy, in the sense that the less ambiguous paths are validated first, thereby making the solution reasonably independent of the order in which nodes are scheduled as key-node and appearance hypothesis are tested;
\item The second one consists in defining the size of the key-node neighborhood proportionally to the length of the key-node. This makes the aggregation multi-scale, which benefits both the accuracy and the computational efficiency, since the individual detections get the opportunity to be aggregated into tracklets before investigating large time horizons, leading to less nodes and more accurate appearance estimation on large time frames. See Section \ref{section:multiscale_iterative_hypothesis_testing}.
\end{itemize}

The global flow of our proposed iterative aggregation algorithm is presented in Algorithm~\ref{algo:IHT}. 
\begin{algorithm}[h]
	\caption{Iterative Hypothesis Testing}
	\label{algo:IHT}	
        \begin{algorithmic}
		\REQUIRE{Graph $\cl G=(\cl V,\cl E,\bs W)$, number of iterations \texttt{MAX\_ITER}}
		\ENSURE {Updated graph after \texttt{MAX\_ITER} iterations}
	\end{algorithmic}
        \textbf{Procedure:}
	\begin{algorithmic}
		\STATE{$dir \leftarrow +1$} 
		\STATE{Initialize $K_1^{(1)}$ and $K_2^{(1)}$}
		\FOR{$l=1,\cdots,$ \texttt{MAX\_ITER}}
			\STATE{{Initialize:} $\mathcal{R} \leftarrow \mathcal{V}$}
			\WHILE{$\mathcal{R} \neq \emptyset$}
				\STATE{$v_{\rm key} \leftarrow$ {Schedule}($\mathcal{R}$)}
				\STATE{$v_{\rm agg} \leftarrow $ {HypothesisTesting}($\cl G,v_{\rm key},dir, K_1^{(l)}, K_2^{(l)}$)}
				\IF{$v_{\rm agg} \neq v_{\rm key}$}
					\STATE{$\mathcal{G} \leftarrow $ {Simplify}$\left(\mathcal{G},v_{\rm agg}\right)$}
				\ENDIF				
				\STATE{$\mathcal{R} \leftarrow \mathcal{R}\setminus {v_{\rm agg}}$}
			\ENDWHILE
			
			\STATE{$(K_1^{(l+1)},K_2^{(l+1)}) \leftarrow \text{Relax}(K_1^{(l)},K_2^{(l)})$}
			\STATE{$dir \leftarrow -dir$}
		\ENDFOR
	\end{algorithmic}
\end{algorithm}

As controlled by the $dir$ flag, the direction of investigation changes
at each graph-scanning iteration to propagate the key-node appearance hypothesis both forward and backward, thereby making the global process symmetric with respect to time. 

In Algorithm~\ref{algo:IHT}, the function \textbf{Schedule} selects a node for hypothesis testing that has not yet been scheduled during the on-going scanning of the graph. In this paper, we select the nodes in decreasing order of their lengths because long nodes are more likely to have accumulated reliable appearance information. Our experimental results have shown that the node scheduling strategy does not affect the performance much.

The remainder of this section details the practical implementation of the core of our proposed \textbf{HypothesisTesting} strategy. It is detailed in Algorithm~\ref{algo:hypothesis_testing} and involves both (i) the computation of the shortest-path connecting the key-node to its neighborhood, under target appearance hypothesis, and (ii) the validation or rejection of this path as a tracklet for subsequent iterations of Algorithm~\ref{algo:IHT}.
\subsubsection{Hypothesis: multi-scale tracklet aggregation}
\label{section:multiscale_iterative_hypothesis_testing}
Formally, the key-node is denoted $v_{\rm key}$. It is selected among the set of nodes, $\mathcal{R}$, that have not yet been investigated during the current scanning of the graph. The aggregation of the key-node with its neighbors is then investigated in an observation window that precedes or follows the key-node, depending on the sign of the \emph{dir} flag. The size of the observation window is proportional to the length of the key-node. 
We use $\delta$ to denote the observation window interval and $|\delta|$ to denote its size. Hence, $\delta = [t_{v_{\rm key}}^{(e)}, t_{v_{\rm key}}^{(e)} + \kappa \cdot |v_{\rm key}|]$ in the forward mode ($dir=1$), or ${\delta = [t_{v_{\rm key}}^{(s)} - \kappa \cdot |v_{\rm key}|, t_{v_{\rm key}}^{(s)}]}$ in the backward mode ($dir=-1$), where $\kappa\in \mathbb{R}_+$ is the window proportionality constant. 

\begin{algorithm}[h]
	\caption{\textbf{HypothesisTesting}}
	\label{algo:hypothesis_testing}	
    \begin{algorithmic}
		\REQUIRE{Graph $\cl G$, key-node $v_{key}$, direction flag $dir$, validation parameters $K_1, K_2$}
		\ENSURE {Nodes that can be aggregated $v_{agg}$}
	\end{algorithmic}
        \textbf{Procedure:}
	\begin{algorithmic}
		\STATE{$\delta \leftarrow$ Limits of observation window \{\emph{See text}\}}
		\STATE{$\cl G_{\delta} \leftarrow \text{GraphHypothesis}(\cl G,\delta,v_{\rm key})$}
		\STATE{$(S_b,S_{sb}) \leftarrow $ Shortest- and second shortest-paths from $v_{\rm key}$}
		\IF[\emph{Refer to Figure~\ref{Fig:graph_observation_window} for illustration} ]{\texttt{isUnambiguous}($S_b, S_{\rm sb}$)} 
			\STATE{$\cl G^{-}_{\delta} \leftarrow $ {ReverseDirection}($\cl G_{\delta}$)}
			\STATE{$(S_{b^\prime},S_{sb^\prime}) \leftarrow $ Shortest- and second shortest-paths from $v_{b}$}
			\IF{{isUnambiguous}($S_{b^\prime}, S_{\rm sb^\prime}$)}
				\STATE{$v_{agg} \leftarrow S_b$}
			\ENDIF
		\ELSE
			\STATE{$v_{agg} \leftarrow v_{key}$}
		\ENDIF	
		\RETURN{$v_{agg}$}
	\end{algorithmic}
	\vspace{2mm}
		\underline{{isUnambiguous}($S_b, S_{\rm sb}$)}
		\begin{algorithmic}
			\RETURN {$\text{cost}(S_b) < K_1\cdot|\delta|$ and $\text{cost}(S_b)/\text{cost}(S_{sb}) < K_2$}
		\end{algorithmic}		
\end{algorithm}

Given the key-node $v_{\rm key}$ and the observation window $\delta$, we define a graph $\mathcal{G}_{\delta}$ to investigate how the key-node can be aggregated with its neighbors to define an appearance-consistent path under the assumption that the target appearance is defined by the key-node appearance. 
The graph $\mathcal{G}_{\delta}$ is directly derived from the graph $\cl G$, by cutting $\cl G$ according the limits of the observation window, and updating the inner costs of the nodes within the window to reflect the hypothesis made about the target appearance. In short, the inner cost $w_{v}$ of a node $v \in \cl V_{\delta}$ is increased (decreased) if it has a different (similar) appearance than the one of the key-node. 

In more details, the inner cost is updated as follows. First, an appearance is associated to the tracklet $v$. The inference of the tracklet appearance from its individual detections appearances directly depends on the characteristics of the appearance observation process. If, for example,  the observation process is affected by outliers, a RANSAC \mycite{fischler1981random} approach would be appropriate to capture the right tracklet appearance. In contrast, if the observations are independent and affected by Gaussian noise, then a weighted average provides an appropriate inference mechanism. Here, we use a weighted average for the tracklet appearance as an example of possible practical implementation. 
Then, the average $i^{\rm {th}}$ feature of a node $v$ is computed as

\begin{equation}
  \overline{\bs f}_i^{(v)}=\frac{1}{C_i^{(v)}}
  \sum_{t=1}^{|v|}{{c}_{i,t}^{(v)} {\bs f}_{i,t}^{(v)}},
    \label{equation:v_a}
\end{equation}
where ${C_i^{(v)}}=\sum_{t=1}^{|v|}{c}_{i,t}^{(v)}$. 

Given the key-node and tracklet $v$ appearances $\overline{\bs f}_i^{(\rm key)}$ and $\overline{\bs f}_i^{(v)}$ respectively, let ${D}(v)$ denote the value by which the inner cost of node $v$ is incremented due to its dissimilarity with respect to the key-node appearance. We define 

\begin{align}
		D(v) = \sum_{i=1}^{\numfeat} \underbrace{\bigl[ \alpha_i^{(\rm key)}\alpha_i^{(v)} \lambda_i \big\|\overline{\bs f}_i^{\rm (key)}-\overline{\bs f}_i^{(v)}\big\|_1+(1-\alpha_i^{(\rm key)}\alpha_i^{(v)}) w_i^{(\rm fix)}  \bigr]}_{w_i^{(v)}},
	\label{equation:appearance_cost}	
\end{align}

where $\lambda_i$ weights the contribution of the $i$-th feature. 
The parameter $\alpha_i^{(v)}$ is introduced such that it tends to one (zero) for confident (unreliable) features. 
As an example of practical implementation, we define it as:
\begin{equation}
	\alpha_i^{(v)}=\begin{cases}
				0 & {\mbox{if } C_i^{(v)} \leq C_{\rm min}, }\\
				1 & {\mbox{if } C_i^{(v)} \geq C_{\rm max},}  \\
				\frac{C_i^{(v)}-C_{\rm min}}{C_{\rm max}-C_{\rm min}} & {\mbox{otherwise.}}
			\end{cases}
	\label{equation:reliability}
\end{equation}
where $C_{\rm min}$ and $C_{\rm max}$ are the limits to define if the feature is considered reliable or not.

From Equation~\ref{equation:appearance_cost}, when $\alpha_i^{(\rm key)}\alpha_i^{(v)} \rightarrow 1$, $w_i^{(v)} \rightarrow \lambda_i \big\| \overline{\bs f}_i^{\rm (key)}-\overline{\bs f}_i^{(v)} \big\|_1$ and when $\alpha_i^{(\rm key)}\alpha_i^{(v)} \rightarrow 0$, $w_i^{(v)} \rightarrow w_i^{(\rm fix)}$.
The term $w_i^{(\rm fix)}$ is introduced so that a node that definitely looks similar to the key-node (${D}(v) \approx 0$) is favored compared to a node for which no appearance features is available $\big({D}(v) \approx \sum_i w_i^{(\rm fix)}\big)$. Empirically, we set ${w_i^{(\rm fix)}=5}$ for all ${1\leq i\leq \numfeat}$.
  
After the  inner costs of the nodes have been incremented by ${D}(v)$, the shortest-path $S_b$ to connect the key-node to the extremity of the observation window is computed. 
Thanks to the directed and acyclic nature of the graph, the shortest-path computation can exploit the inherent topological ordering of the nodes (\eg, to support a depth-first search) which is more efficient that the Dijkstra's algorithm. The cost of a path is defined to be the sum of costs of the edges and the inner costs of the nodes along it, and is given by the function \textbf{cost} in Algorithm \ref{algo:hypothesis_testing}.

Even though it seems that updating the costs requires additional scanning of the graph, it is mitigated by the concept of \emph{visitors} in the shortest-path algorithm of the Boost Graph Library. The visitors allow to update the costs of the nodes or edges on-the-fly as they are manipulated during the shortest-path computation.
\subsubsection{Testing: path ambiguity estimation and tracklet validation}
\label{section:ambiguity_estimation_and_validation}

Since the cost of the edges have been defined to take the displacement as well as the appearance into consideration, the shortest-path $S_b$, which connects the key-node to the extremity of the observation window, reasonably corresponds to a single physical object (same appearance, and consistent displacements) and could thus be aggregated into a single node. 

\begin{figure}[ht]
	\centering
	\includegraphics[width=0.5\linewidth]{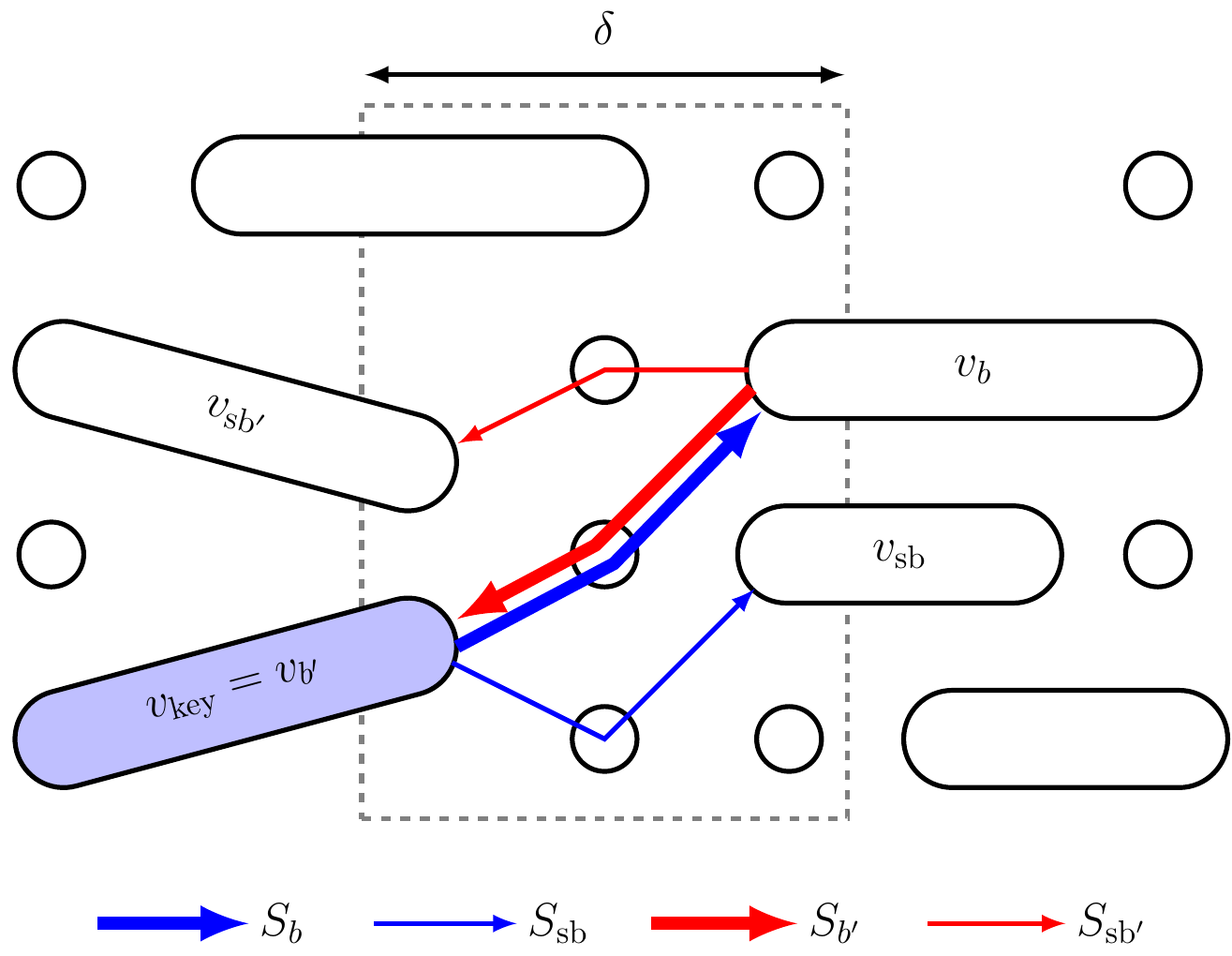}
	\caption[Illustration of the validation of the hypothesis]{\textbf{Illustration of the validation of the hypothesis}. Within the window, the best (thick arrow) and the second best (thin arrow) paths (denoted by $S_b$ and $S_{\rm sb}$ respectively) are searched. Blue and red arrows represent forward and backward directions respectively. \textit{ Best viewed in color.}}
	\label{Fig:graph_observation_window}
\end{figure}

However, to limit the risk of connecting nodes that correspond to two distinct objects, we check the level of ambiguity of the shortest-path by comparing its cost to the costs of alternative paths. Figure~\ref{Fig:graph_observation_window} illustrates this process.

It runs in two steps. In the first step, the shortest $S_b$ and the second shortest $S_{\rm sb}$ paths\footnote{In our implementation, the second best path $S_{\rm sb}$ is chosen to be a path that does not overlap the shortest-path $S_b$. When there exist paths with smaller costs that partly overlap with $S_b$, the unshared part of $S_b$ remains subject to ambiguity, even if $S_{\rm sb}$ is large. Hence, only the part of $S_b$ that is shared with those alternative paths should be considered for aggregation. For details, refer to the reference software at \urlispg.} are considered. The ends of the best and second-best paths are denoted as $v_{\rm b}$ and $v_{\rm sb}$ respectively. The shortest-path $S_b$ is considered being unambiguous  only if two conditions are met: (i) cost\big($S_{\rm b}$\big)$<K_1 \cdot |\delta|$, and (ii) cost\big($S_{b}$\big)/cost\big($S_{\rm sb}$\big)$<K_2$. 

If all conditions are met, the second step of the validation process is considered. For this, the graph is reversed by flipping the direction of all the edges of $\cl G_{\delta}$. It is mentioned as \textbf{ReverseDirection} in the algorithm. The shortest- ($S_{b'}$) and second shortest- ($S_{\rm sb'}$) paths linking $v_{\rm b}$ with the opposite extremity of the observation window are then computed. If $S_{b'}$ leads to the original key-node, {\em i.e.}, if $v_{\rm b'}=v_{\rm key}$, and if a similar set of conditions 
hold for $S_{b'}$ and $S_{\rm sb'}$, then the path $S_{b}$ is considered to be \textit{unambiguous}, and is replaced by a single node in the graph for subsequent iterations of the IHT. This procedure is called \textbf{Simplify} in the Algorithm~\ref{algo:IHT}. It updates the appearance features of the node as in Equation~\ref{equation:v_a} and also the motion parameters. It keeps only the edges connecting the extremities of the aggregated path to the rest of the graph. Other connections involving intermediate nodes are removed. 


Choosing small (large) values of $K_1$ and $K_2$ makes the constraint more (less) conservative. In the first iterations of the algorithm, we start with small values of $K_1$ and $K_2$. As the iterations proceed, we progressively relax the validation criteria. This makes the overall IHT algorithm greedy, in the sense that the less ambiguous paths are validated first, thereby making the solution reasonably independent of the order in which nodes are scheduled as key-node and appearance hypotheses are tested. This progressive relaxation of the key-node path validation constraint is denoted by the function \textbf{Relax} in Algorithm~\ref{algo:IHT}. An example of relaxation scheme is described in results section. 
\section{From off-line to incremental IHT}
\label{section:incremental_iht}
Because we iterate over the nodes, our IHT naturally extends to the incremental scenarios in which the detections arrive sequentially over time. Compared to the off-line approach, there are however few subtleties. They are:
\begin{itemize}
	\item \textbf{Incrementing the graph:} At time $t=1$, the graph is just a set of detections at that instant, \ie, $\cl G^1=(\cl D^1, \emptyset)$. At time $t>1$, the graph is obtained by adding new detections $\cl D^t$ to the so-called previous graph $\cl G^{t-1}$, resulting from earlier steps of the algorithm, up to time $t-1$. All nodes ending later than time $t-\tau_{\rm max}$ are linked to all the current detections. The weight of each edge is computed as in Equation~\ref{eqn:spatio_temporal_cost}.
	\item \textbf{Scheduling of the nodes:} Unlike the off-line approach, we schedule the `recent' nodes first. This is done to prevent the fast growth of the graph at each time. Specifically, we schedule the nodes in decreasing order of $|v|/\max\{1,t-t_v^{(e)}\}$ so that the `recent' and `sufficiently long' nodes are selected first.
	\item \textbf{Relaxing the validation criteria:} We maintain a `sliding window' $[t-\delta_{\rm slide},t]$ where $\delta_{\rm slide}$ is the length of the sliding window. Inside (respectively, outside) the sliding window, we impose conservative (respectively, relaxed) criteria for $K_1$ and $K_2$. We use $\delta_{\rm slide}=200$ frames.
\end{itemize}
\section{Evaluation}
\label{section:evaluation_iht}
We test our proposed IHT algorithm on a toy example and also on the real-life APIDIS \mycite{apidis}, PETS \mycite{pets2009_dataset} and TUD \cite{tud_dataset} datasets. The toy example helps us to highlight the various steps of our IHT aggregation paradigm, while the experiments on real-life examples demonstrate the practical relevance of our approach.

The proposed approach has been implemented in C++ (for APIDIS dataset) and MATLAB\footnote{The MATLAB implementation is available at the \urlispg.} (for toy example and PETS dataset). The C++ implementation utilizes Boost Graph Library for representing the graph. The DAG shortest path algorithm is provided in the library. All experiments are performed on a desktop computer with 3GHz
quad-core CPU, 4 GB of RAM, and running under Linux. 

\subsection{Evaluation metrics}
\label{section:evaluation_metric}
We use the CLEAR MOT metric \cite{clear_mot} to evaluate our approach. It defines two quantities, namely the multiple object tracking precision (MOTP) and the multiple object tracking accuracy (MOTA).

MOTP is defined as the average error in estimated position of pairs of matched ground-truth and estimated track pairs. MOTA is defined to decrease proportionally to the number of missed detections, false positives, reinitializations and switches (see \cite{clear_mot} for the formal definition). 
The error due to switches is usually more problematic since it affects the higher level interpretation of the tracks.

Since MOTP depends on the accuracy of target detector and on the accuracy of ground-truth accuracy, MOTA is often preferred over MOTP. Since switching errors are important, we also report the overall switching errors (SW).

\subsection{Datasets}
\label{section:datasets}
\textbf{Toy dataset.} This dataset is considered to observe how our algorithm compares to related works, but also to assess its sensitivity to parameters selection. We consider 3 targets whose ground-truth locations $\{y_1, y_2, y_3\}$ at time instances $k \in \{0,\cdots,10\}$ are obtained by
\begin{align}
	y_1:=50 \sin\left(\frac{2\pi k}{10}\right),\; y_2:=50 \cos\left(\frac{2\pi k}{10}\right), \; y_3&=-20-50 \sin\left(\frac{2\pi k}{8}\right)
	\label{eqn:toy_gt_location}
\end{align}

\begin{figure}[h]
  \centering  
  \includegraphics[width=0.5\linewidth]{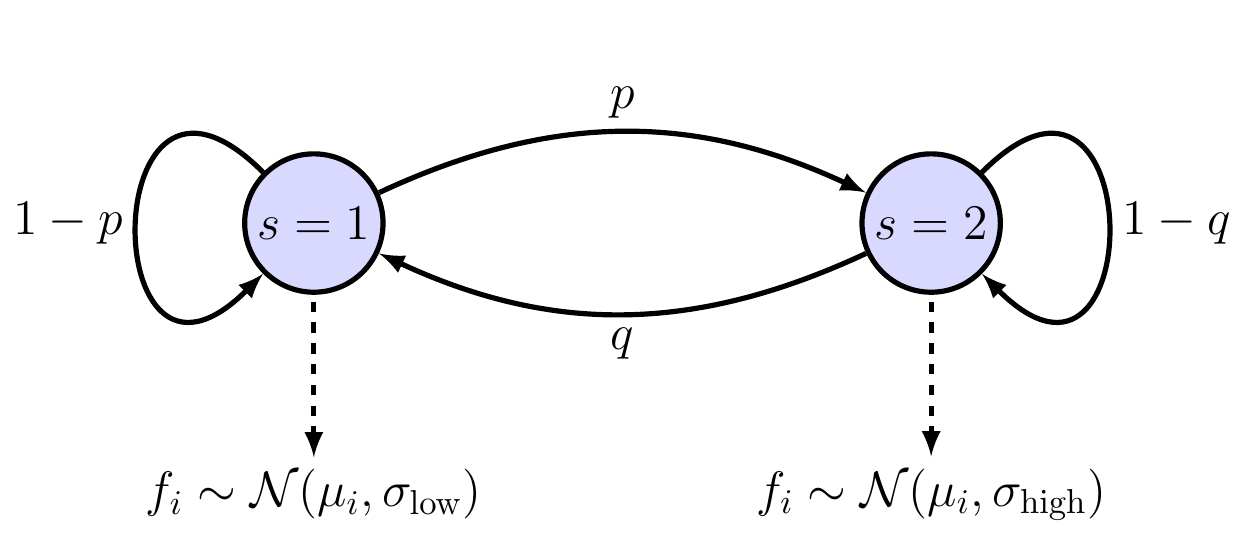}
  \caption[2 state automata for modelling the appearance of the $i$-th target]{\textbf{2 state automata for modelling the appearance of the $i$-th target}.}
  \label{fig:markov_chain}
\end{figure}

The appearance feature of the $i$-th target, denoted as $f_i$, is modeled by a 2 state automata as shown in Figure\ref{fig:markov_chain}. 
For the $i$-th target, the appearances of the state 1 and 2 are modelled as $\mathcal N(\mu_i,\sigma_{\rm low})$ and $\mathcal N(\mu_i,\sigma_{\rm high})$ respectively. We use $\mu_i \in \{0,120,240\}$ and $\sigma_{\rm low}=10$ and $\sigma_{\rm high}=100$. 
We fix $q=0.5$ and vary $p$. The confidence of the feature measurement process is estimated as $c_i=0.1$ if $s=2$, and 0.8 if $s=1$.


\textbf{APIDIS dataset.} This 15 minutes long basketball video dataset \cite{apidis} has been captured by 7 cameras. The candidate detections are computed at each time instant based on a ground occupancy map, as described in \cite{delannay2009detection}. For each detection, jersey color and digit are considered appearance features. The jersey color is computed as the average blue component divided
by the sum of average red and green components,
over the foreground silhouette of the player within
the detected rectangular box. The digit feature is
obtained by running a digit-recognition algorithm \cite{verleysen2012recognition}
in the same rectangular region. The digit feature is
inherently sporadic as it is available only when the
digit faces the camera. 

\textbf{Pedestrian datasets.} We use publicly available PETS S2/L1 and TUD Stadtmitte datasets to evaluate the performance of our approach on monocular views. The PETS is a 795 frames long dataset with moderate target density. TUD Stadtmitte is 179 frames long. Because of the low view-point, the targets frequently occlude each others. Detection results are obtained from \cite{anton_website}. At each detection, we compute 24-bin color histogram by concatenating 8-bin RGB color histograms\footnote{In a tracklet, a distinct histogram is associated to each extremity, so as to account for progressive target appearance changes.}. We ignore the color histogram if the overlap ratio between two bounding boxes exceeds 10\%. This is done as the histograms are likely to be unreliable in presence of occlusions. 


\subsection{Results on the toy example}
For the toy example, the nodes of the graph $\cl G$ correspond to the detections defined by Equation (\ref{eqn:toy_gt_location}). For IHT, KSP and GAC, we create edges between the nodes that occur at consecutive time instants. The cost $w_{ij}$ writes
%
%
		$w_{ij}:=w^{(s)}_{ij}+w^{(a)}_{ij}$, where $w^{(s)}_{ij}$ and $w^{(a)}_{ij}$ are the spatio-temporal and appearance costs respectively. The spatio-temporal cost  is defined as $w^{(s)}_{ij}:=\|y_j-y_i\|_2$. The appearance cost differs from one algorithm to another. Given two appearance features $f_i$ and $f_j$, the appearance dissimilarity $d_{ij}$ is computed as $d_{ij}:=1-|\cos(\pi(f_j-f_i)/180)|$. Then, we define the appearance cost $w^{(a)}_{ij}$ as in Equation~(\ref{equation:appearance_cost}):
		\begin{center}
			\begin{tabular}{ccc}
				\hline
				\textbf{Algorithm} & \textbf{Reference appearance} & \textbf{Appearance cost}, $w_{ij}^{(a)}$\\
				\hline
				\textbf{KSP} & None & $c_i c_j d_{ij}+(1-c_ic_j)w^{(\rm fix)}$\\
				\textbf{GAC} & $l$-th global appearance, $f_l$ & $c_i d_{il}+(1-c_i)w^{(\rm fix)}$\\
				\textbf{IHT} & $l$-th key-node appearance, $f_l$ & $c_i c_l d_{il}+(1-c_ic_l)w^{(\rm fix)}$\\
				\hline
			\end{tabular}
		\end{center}
		where $w^{(\rm fix)} \geq 0$ is a fixed cost, introduced to associate a fixed cost to nodes for which the appearance is unknown or unreliable. We use $w^{(\rm fix)}=10$ in our experiments. 
		For GAC, the set of global appearances considered by GAC are either known \emph{a priori} (\eg, provided by oracle), or are estimated from the measurements (\eg, using k-means with 3 clusters in our toy-example case).	
	For GMCP, we consider the problem in Equation \ref{eqn:generic_definition_cost_track} with $C(T_i):=\sum_{u \in T_i} \bigl[ \sum_{\substack{v \in T_i \\ v \neq u}} w_{uv}^{(a)}+w_{uv^\star}^{(s)}\bigr]$, where $v^\star=\arg\!\min_{\substack{v \in T_i \\ v \neq u \\t_v >t_u}} (t_v-t_u)$. We define $w_{uv}^{(s)}:=\|y_u-u_{v^{\star}}\|_2$ if $v=v^\star$ and $w_{uv}^{(s)}:=\infty$ otherwise. The appearance cost $w_{uv}^{(a)}$ is defined similarly to KSP. As told in Section~\ref{section::previous_art_simplification_iht}, this GMCP formulation is NP-complete. We have followed the authors in \mycite{gmcp_tracker}, and have adopted the popular 2-opt local search \mycite{croes1958method} to solve it. 
		
\begin{figure}[h]
	\centering
	\includegraphics[width=0.9\linewidth]{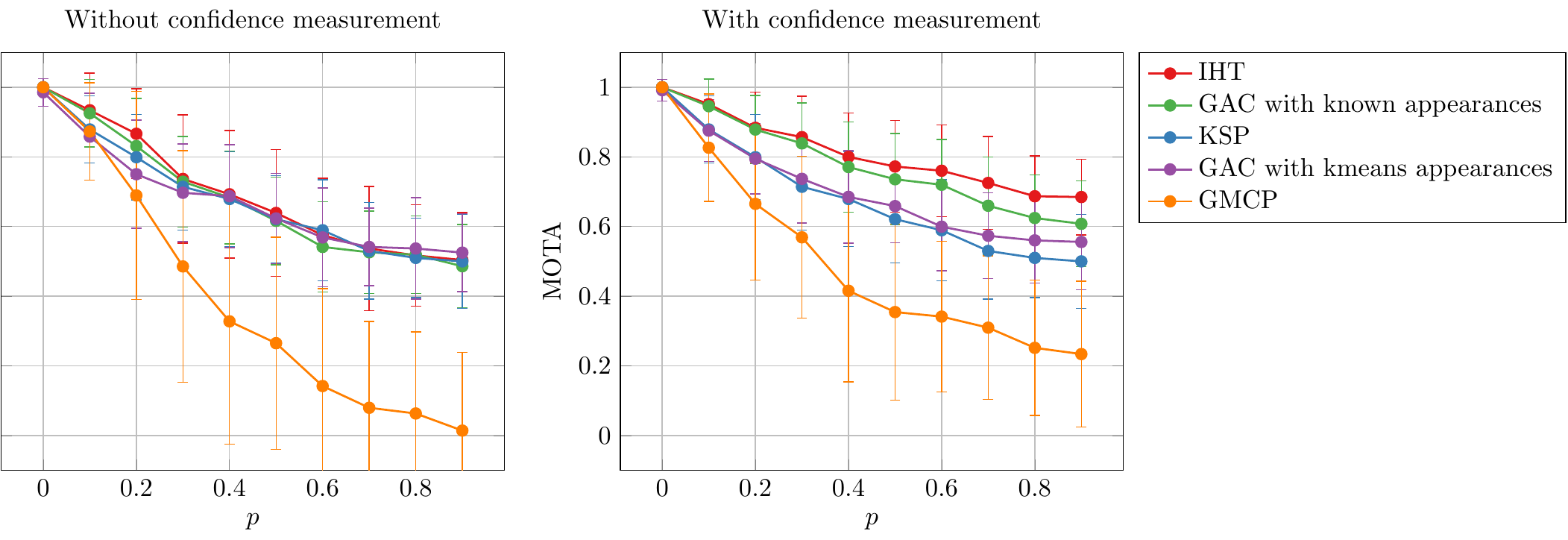}
	\caption{\textbf{Performance of IHT, GAC, KSP and GMCP on the toy example} with and without taking the feature measurement confidence information. \textit{ Best viewed in color.}}
	\label{fig:error_bar_gac_iht}
\end{figure}

\textbf{Results:}
In our simulations, we vary the transition probability $p$ from $0$ to $0.9$ with an increment of $0.1$. For each value of $p$, we generate 100 realizations of the target appearances and apply IHT, GMCP, GAC and KSP algorithms. 
The MOTA obtained with and without prior knowledge about appearance measurement reliability, \ie,  with and without knowing the state of the automata, are shown in Figure~\ref{fig:error_bar_gac_iht}.

We observe that taking the confidence of the feature measurement into account helps to disambiguate the data association. When we do not take into account the confidence information, all IHT, GAC and KSP perform similarly, with IHT performing slightly better than the other two algorithms. The  performance improves significantly when the confidence measure is incorporated. Surprisingly, the GMCP has the worst performances even though it adopts a close to ideal problem formulation. The inferior performance of GMCP can be accredited to the fact that each `track' is extracted greedily and locally  from the set of nodes. Unlike GAC, there is no notion of global solution. Unlike IHT, it does not challenge the ambiguity of the extracted track.

It is worth noting that the performance of GAC is strongly dependent on the prior knowledge of the 3 global appearances. The performance in Figure~\ref{fig:error_bar_gac_iht} indeed appears to degrade significantly when the 3 appearances are estimated from the measurements (based on k-means clustering).

\begin{figure}[h]
	\centering
	\includegraphics[width=0.6\linewidth]{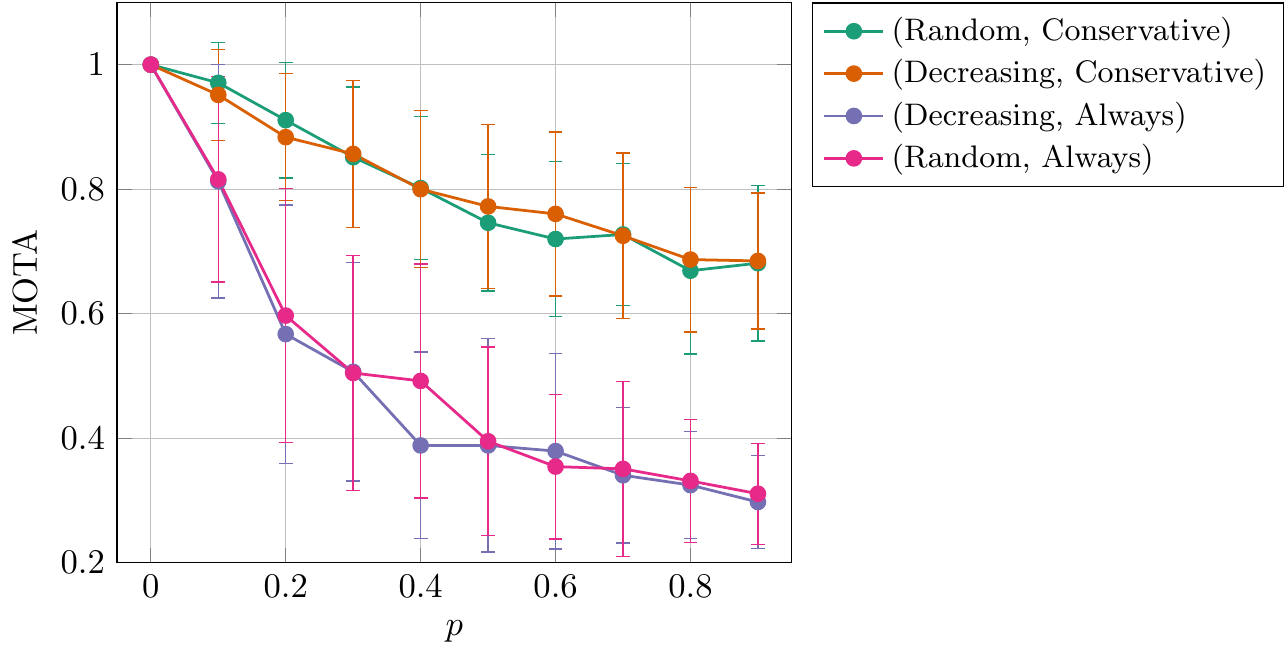}
		\caption[Effect of scheduling of nodes and validation strategy on the performance of IHT]{\textbf{Effect of scheduling of nodes and validation strategy on the performance of IHT}. 
		\textit{Best viewed in color.} }
		\label{fig:performance_iht_schedule_validate}
\end{figure}

Our IHT algorithm has two distinct steps: (i) node scheduling, and (ii) hypothesis validation. To study the importance of these steps, we envision the following set-ups. We schedule the nodes either at \emph{random} or in \emph{decreasing} order of appearance confidence. In addition, we validate the shortest-path either \emph{conservatively}, as described by Figure~\ref{Fig:graph_observation_window}, or \emph{always}, meaning that we systematically define a new tracklet based on the shortest-path. The results are presented in Figure~\ref{fig:performance_iht_schedule_validate}. They show that the node scheduling has negligible impact on the performance of the IHT algorithm. On the other hand, the conservative validation of the shortest path has a drastic influence on the performance of IHT. By comparing Figure~\ref{fig:performance_iht_schedule_validate} with Figure~\ref{fig:error_bar_gac_iht}, we observe that IHT performs worse than KSP when we validate the shortest path immediately. This is not surprising because IHT investigates on a local section of the graph whereas KSP works globally on the whole graph. 

\subsection{Results on real-life datasets}
\label{section:results_offline_iht_apidis}
In this section, we present and discuss the performances of both the off-line and incremental IHT. To better compare our method with the literature, we consider two experimental set-ups. The first one discards the appearance features and uses only the spatio-temporal information. In contrast, the second one incorporates the appearance features.

Apart from KSP, GAC and GMCP, we compare our results with several other methods such as the discriminative label propagation (DLP) \cite{kc2013discriminative} (introduced in Section \ref{section:related_works_iht}), the continuous energy (CE) \cite{anton_continuous_minimization}, the discrete-continuous optimization (D-C) \cite{discrete_continuous}. The CE and D-C trackers compute the most probable tracks by minimizing a combination of energies that reflect the consistency with the observed detections locations, the plausibility of the tracks dynamics, the persistence of tracks, and the exclusivity constraint between co-existing tracks. In addition, D-C uses cubic splines to model the motion of targets and also penalizes the number of trajectories. 

It is to be noted that CE and D-C do not use appearance features. Therefore, we compare them with the first version of IHT that does not exploit appearance features. GAC and DLP are able to exploit sporadic appearance features only. GMCP, on the other hand, can exploit appearance features that can be sporadic or have variable reliability. Therefore, we compare the second version of our IHT with these methods.

\begin{table}[h]
\centering
\begin{tabular}{llccc|ccc}
\multicolumn{1}{l}{\multirow{2}{*}{{\bf Dataset}}} & \multirow{2}{*}{{\bf Method}} & \multicolumn{3}{c|}{{\bf No appearance}}                                                         & \multicolumn{3}{c}{{\bf With appearance}}                                                     \\ \cline{3-8} 
\multicolumn{1}{c}{}                               &                               & \multicolumn{1}{c}{{\bf MOTA}} & \multicolumn{1}{c}{{\bf MOTP}} & \multicolumn{1}{c|}{{\bf SW}} & \multicolumn{1}{c}{{\bf MOTA}} & \multicolumn{1}{c}{{\bf MOTP}} & \multicolumn{1}{c}{{\bf SW}} \\ \hline
\multirow{4}{*}{APIDIS}                            & GAC \cite{shitrit2011global}                          & 72.91                          & 53.13                          & 108                           & 73.07                          & 53.15                          & 110                          \\
                                                   & DLP  \cite{kc2013discriminative}                         & 81.25                          & 57.13                          & 49                            & 83.80                          & 60.01                          & 45                           \\ \cline{2-8} 
                                                   & IHT (offline)                 & 76.71                          & 65.36                          & 11                            & 87.91                          & 64.43                          & 0                            \\
                                                   & IHT (incremental)             & 75.99                          & 64.53                          & 14                            & 86.82                          & 65.13                          & 3                            \\ \hline
\multirow{6}{*}{TUD}                    & CE  \cite{anton_continuous_minimization}                          & 60.5                           & 65.8                           & 7                             & -                              & -                              & -                            \\
                                                   & D-C  \cite{discrete_continuous}                         & 61.8                           & 63.2                           & 4                             & -                              & -                              & -                            \\
                                                   & GMCP  \cite{gmcp_tracker}                        & -                              & -                              & -                             & 77.7                           & 63.4                           & 0                            \\
                                                   & DLP \cite{kc2013discriminative}                          & 62.6                           & 73.5                           & 17                            & 79.3                           & 73.9                           & 4                            \\ \cline{2-8} 
& IHT (offline) & \revise{62.1} & \revise{73.2} & \revise{7} & \revise{78.5} & \revise{73.2} & \revise{0} \\
& IHT (incremental) & \revise{61.8} & \revise{72.9} & \revise{9} & \revise{78.3} & \revise{73.1} & \revise{1} \\ 
\hline
\multirow{7}{*}{PETS}                              & CE \cite{anton_continuous_minimization}                           & 81.84                          & 73.93                          & 15                            & -                              & -                              & -                            \\
                                                   & D-C \cite{discrete_continuous}                          & 89.30                          & 56.40                          & -                             & -                              & -                              & -                            \\
                                                   & GMCP \cite{gmcp_tracker}                         & -                              & -                              & -                             & 90.30                          & 69.02                          & 8                            \\
                                                   & DLP  \cite{kc2013discriminative}                         & 82.75                          & 71.21                          & 25                            & 91.01                          & 70.99                          & 5                            \\
                                                   & GAC \cite{shitrit2011global}                          & 80.00                          & 58.00                          & 28                            & 81.46                          & 58.38                          & 19                           \\ \cline{2-8} 
& IHT (offline)  &  \revise{81.18} & \revise{74.53} & \revise{9} & \revise{85.10}  & \revise{74.56}  & \revise{4}  \\
& IHT (incremental)   & \revise{80.91}  & \revise{74.48}  &  \revise{11}  & \revise{84.78} & \revise{74.32} & \revise{5} \\ 
\hline
\end{tabular}
\caption{Tracking results on the APIDIS (1500 frames), PETS (795 frames) and TUD Stadtmitte (179 frames) datasets.}
\label{table:tracking_results_iht}
\end{table}

From Table \ref{table:tracking_results_iht}, we first observe that the incremental version performs slightly worse than the off-line version. For APIDIS dataset, our method outperforms KSP and GAC significantly. Even though DLP seems to work better than IHT when no appearance features are used, it commits significant switching errors. This illustrates the conservativeness of IHT. When the appearance features are incorporated, IHT outperforms all other methods.

In case of pedestrian datasets, IHT seems to perform comparably with other methods. Even though the MOTA scores are similar or  lower than GMCP and DLP, the number of switching errors are significantly lower for IHT, which is an advantage in terms of high-level scene interpretation. In case of TUD dataset, IHT is comparable to CE, D-C and DLP in terms of MOTA. However, in case of PETS dataset, IHT performs worse than D-C. The superior performance of D-C in scenarios for which no appearance feature is exploited can be accredited to the fact that this approach use higher order motion models. 
On the positive side, our tracker commits fewer switching error.

\begin{figure}[h]
	\centering
	\includegraphics[width=0.9\linewidth]{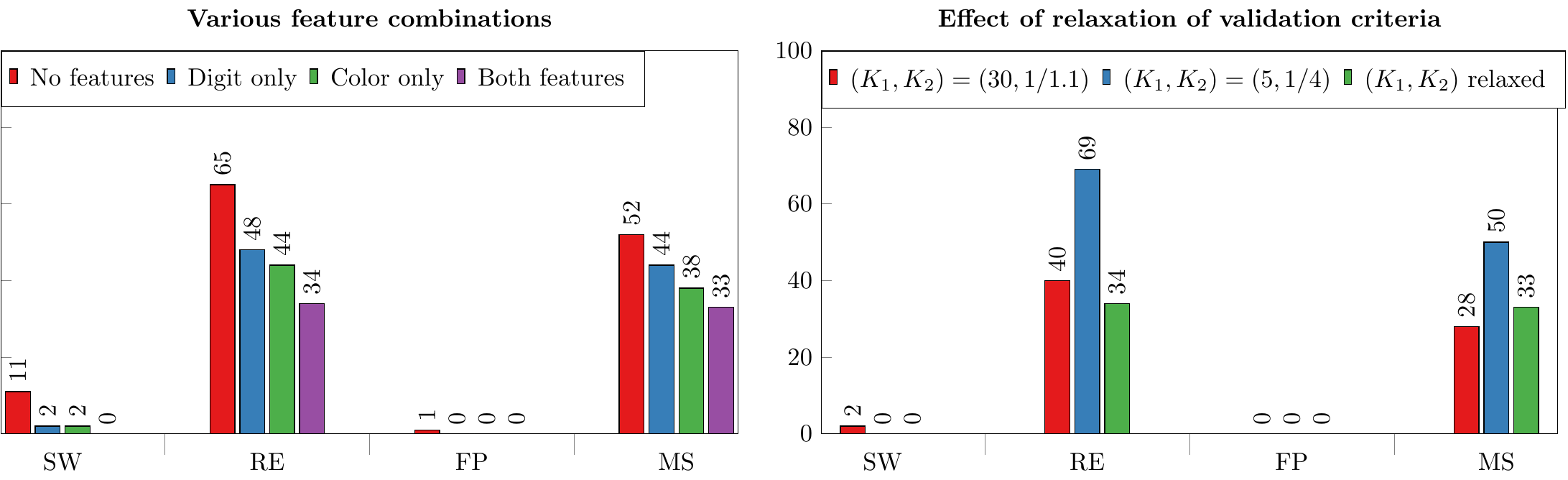}
	\caption{\textbf{Components of MOTA metric on a 1 minute long video of the APIDIS dataset for off-line IHT.} \textbf{(Left.)} Various feature combinations. The MOTA scores for all 4 cases are as follows: (i) no feature: 76.71\%, (ii) digit feature only: 83.03\%, (iii) color feature only: 84.84\%,  and (iv) both features: 87.91\%. \textbf{(Right.)} Effect of different $(K_1, K_2)$. Red, blue and green bars correspond to the `least conservative', `most conservative' and `progressively relaxed' validation criteria respectively. The MOTA scores for all 3 cases are as follows: (i) least conservative: 87.36\%, (ii) most conservative: 78.52\%, and (iii) progressively relaxed: 87.91\%. \bestviewed}
	\label{figure:without_and_with_features_offline_iht}
\end{figure}

The right side of Figure~\ref{figure:without_and_with_features_offline_iht} compares the performance obtained by IHT when different sets of appearance features are exploited. 

As we can see, the switches and re-initializations are reduced substantially when the appearance features are used. It can also be seen that the digit features, even though they are highly sparse, can disambiguate some tracks. 

In order to study the effect of the progressive relaxation of the validation criteria, we first fix the values for $K_1$ and $K_2$ such that the validation criteria are `most conservative' (\ie, small values of $K_1$ and $K_2$) and `least conservative' (\ie, large values of $K_1$ and $K_2$). Specifically, we set $(K_1,K_2)=(5,1/4)$ and $(K_1,K_2)=(30,1/1.1)$. For progressive relaxation, we then consider a linear increase in $K_1$ from $5$ to $30$ in $50$ iterations and increase in $K_2$ linearly from $1/4$ to $1/1.1$ in $20$ iterations. 
We increase $K_2$ faster than $K_1$ because the primary condition to validate a path is its low cost. 
The results are depicted in Figure~\ref{figure:without_and_with_features_offline_iht}. We see that relaxing the validation criteria indeed helps to improve the tracking results in the sense that it avoids identity switches (just as for a highly conservative criteria), while maintaining re-initialization and misses at the level obtained with a less conservative criteria. Hence, it keeps the best out of the two criteria.

\begin{figure}[h]
	\centering
	\includegraphics[width=0.9\linewidth]{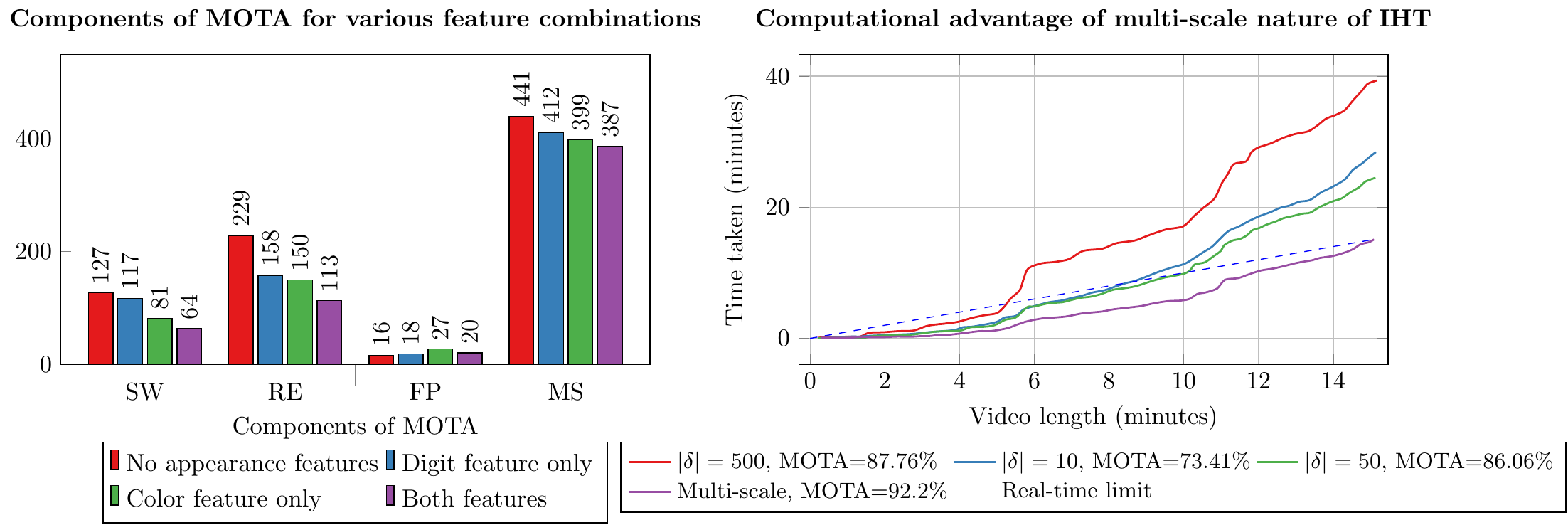}
	\caption{\textbf{Components of MOTA for various feature combinations and computational advantages of multi-scale nature of IHT on a 15 minutes long video.} \textbf{(Left.)}The MOTA scores for all 4 cases are as follows: (i) No appearance: 89.1\%, (ii) Digit only: 90.5\%, (iii) Color only: 91.2\%, and (iv) Both features: 92.2\%. \textbf{(Right.)} Time taken by IHT for multi-scale as well as fixed-window approaches. \textit{ Best viewed in color.}}
	\label{figure:mota_timing_15min}
\end{figure}

To extend our analysis to realistic real-life scenarios (on-the-fly tracking on long sequences), we also report the incremental IHT tracking results for the 15 minutes long APIDIS dataset.
Figure~\ref{figure:mota_timing_15min} compares the performance obtained when different set of appearance features are exploited. 
There are all together 7460 ground truth positions, \emph{i.e.},  GT=7460. As we can see, the switches and re-initializations are reduced substantially when appearance features are exploited. However, the false positives increase slightly. 


To study the computational advantages of our multi-scale approach, we estimate the time taken by IHT with fixed (specifically, $|\delta| \in \{10, 50,500\}$) and adaptive (\ie, $|\delta|=\kappa\cdot |v_{\rm key}|$) observation window sizes. The results are shown in Figure~\ref{figure:mota_timing_15min}. We observe that the multi-scale nature of the algorithm not only reduces the computational time but also improves the tracking accuracy.

We complete our results by presenting the performance of incremental IHT algorithm (in terms of MOTA components) with respect to some key parameters. They are:
\begin{table*}[h]
	\centering
	\small
	\begin{tabular}{ll}
		\textbf{Parameter} & \textbf{Description} \\
		\hline
		$\tau_{\rm max}$ & Connection horizon for new detections (Section \ref{section:graph_construction_iht}, Equation \ref{eqn:spatio_temporal_cost})\\
		$\gamma$ & Missed detection coefficient (Section \ref{section:graph_construction_iht}, Equation \ref{eqn:spatio_temporal_cost})\\
		$\kappa$ & Window proportionality constant (Section \ref{section:multiscale_iterative_hypothesis_testing})\\
		$(C_{\rm min},C_{\rm max})$ & Lower and upper thresholds to compute the reliability (Equation \ref{equation:reliability})\\
		$(K_1,K_2)$ & Factors to validate the shortest-path (Section \ref{section:ambiguity_estimation_and_validation})\\
		\hline
	\end{tabular}
\end{table*}

The reference working point is defined by ($\tau_{\rm max}=120, \gamma=3, \kappa=5, C_{\rm min}=20, C_{\rm max}=100 K_1=5, K_2=1/3$) for which the results are (FP=20, MS=387, RE=113, SW=64, MOTA=92.2\%) for incremental IHT on APIDIS dataset. In Table \ref{table:effect_of_parameters_mota}, only one parameter is changed at a time and all other parameters are fixed at their reference values. 

\begin{table}[h]
\centering
\resizebox{0.99\linewidth}{!}{
\begin{tabular}{cccgc|ccgcc|cgccc}
\multirow{2}{*}{} & \multicolumn{4}{c|}{{$\tau_{\rm max}$}}  & \multicolumn{5}{c|}{{$\gamma$}}             & \multicolumn{5}{c}{{$(C_{\rm min}, C_{\rm max})$}}                       \\ \cline{2-15} 
                  & {\bf 30} & {\bf 60} & {\bf 120} & {\bf 240} & {\bf 1} & {\bf 2} & {\bf 3} & {\bf 4} & {\bf 5} & {\bf (20,70)} & {\bf (20,100)} & {\bf (20,50)} & {\bf (5,50)} & {\bf (20,30)} \\ \hline
{\bf FP}          & 7        & 24       & 20        & 22        & 34      & 20      & 20      & 20      & 11      & 19            & 20             & 19            & 22           & 19            \\
{\bf MS}          & 426      & 410      & 387       & 380       & 429     & 382     & 387     & 382     & 399     & 400           & 387            & 402           & 390          & 402           \\
{\bf RE}          & 162      & 137      & 113       & 111       & 125     & 116     & 113     & 118     & 118     & 127           & 113            & 116           & 133          & 117           \\
{\bf SW}          & 74       & 64       & 64        & 76        & 72      & 83      & 64      & 70      & 71      & 65            & 64             & 70            & 67           & 75            \\ \hline
\end{tabular}
}
\vspace{5mm}
\resizebox{0.99\linewidth}{!}{
\begin{tabular}{cccgc|cgcc|ccgc|c}
\multirow{2}{*}{} & \multicolumn{4}{c|}{$\kappa$}          & \multicolumn{4}{c|}{$K_1$}               & \multicolumn{4}{c|}{$K_2$}                       & \multirow{2}{*}{{\bf Relaxed}} \\ \cline{2-13}
                  & {\bf 1} & {\bf 3} & {\bf 5} & {\bf 7} & {\bf 2} & {\bf 5} & {\bf 15} & {\bf 30} & {\bf 1/1.5} & {\bf 2/2} & {\bf 1/3} & {\bf 1/5} &                                \\ \hline
{\bf FP}          & 14      & 19      & 20      & 33      & 14      & 18      & 24       & 30       & 70          & 32        & 18        & 17        & 20                             \\
{\bf MS}          & 417     & 408     & 387     & 380     & 417     & 401     & 372      & 350      & 363         & 377       & 401       & 445       & 387                            \\
{\bf RE}          & 120     & 126     & 113     & 110     & 133     & 120     & 103      & 97       & 101         & 107       & 120       & 149       & 113                            \\
{\bf SW}          & 62      & 64      & 64      & 73      & 41      & 60      & 69       & 78       & 97          & 81        & 60        & 58        &      64                          \\ \hline
\end{tabular}
}
\caption{Effect of $\tau_{\rm max}, \gamma, \kappa$, $(C_{\rm min},C_{\rm max})$, $K_1$ and $K_2$ on 15 minutes video of APIDIS dataset. For comparison, we also present the results for the case in which $K_1$ and $K_2$ are progressively relaxed.}
\label{table:effect_of_parameters_mota}
\end{table}

From Table \ref{table:effect_of_parameters_mota}, we observe that choosing small $C_{\rm min}$ and  large $C_{\rm max}$ result in a more conservative 
situation, leading to reduced switching errors but increased missed detections. A large connection window $\tau_{\rm max}$ typically is more robust to missed detections but is prone to switching errors. The missed detection penalty $\gamma$ directly affects the missed detection. A small $\gamma$ will creates `short-cuts' in the shortest-path. On the other hand, a big $\gamma$ will favor only temporally local detections. Both situations result in decreased performance. The observation window factor $\kappa$ controls the range in which the key-node appearance hypothesis holds. A small (respectively, large) $\kappa$ investigates small (respectively, large) temporal neighborhood around the key-node. Consequently, small $\kappa$ results in decreased switching errors at the expense of increased misses and re-initializations. Both $K_1$ and $K_2$ affect the performance. Low (respectively, high) values of $K_1$ and $K_2$ result in less (respectively, more) false positives and switching errors but more (respectively, less) misses and re-initializations, allowing us to trade-off the errors. We propose two alternatives to choose these parameters depending on the problem at hand. First, if the objective is to have conservative tracking in which the resulting tracklets are reliable, it is suggested to choose low values of $K_1$ and $K_2$. This option is suitable if one envisions to process these trackets in the next step so as to stitch them into long trajectories. Second, we propose to start with small values of $K_1$ and $K_2$ and then progressively relax as the iteration proceeds. This option is suitable when long (but potentially erroneous) trajectories are preferred. 
\section{Conclusion and future perspectives}
\label{section:conclusion_iht}
This paper proposed a novel framework to associate detections while exploiting {unreliable and/or sporadic appearance features}. It proceeds iteratively, starting with a graph in which each node corresponds to a detection. Each iteration then investigates how to connect a node, named key-node, to its neighbors, under the assumption that the appearance of this key-node is representative of the corresponding target appearance. Unambiguous associations are merged into bigger nodes, thereby creating nodes with more reliable appearance cues. This aggregation also reduces the size of the graphs, and thus the complexity, handled by successive iterations of the algorithm. Defining the size of the neighborhood to be proportional to the size of the key-node naturally ends up in aggregating the data at larger time scales once more appearance cues have been accumulated along the key-node. Progressively relaxing the ambiguity criterion results in a greedy process, that primarily aggregates the less ambiguous paths in the graph.

%
\section*{References}
\bibliography{egbib}

\end{document}